# Chart-Text: A Fully Automated Chart Image Descriptor

**Abhijit Balaji[1], Thuvaarakkesh Ramanathan, Venkateshwarlu Sonathi**

SPi Technologies India Private Limited, Chennai, India
{B.Abhijit, R.Thuvaarakkesh, T.Venkateshwarlu}@spi-global.com

**Abstract**

Images greatly help in understanding, interpreting and visualizing data. Adding textual description to images is the first and foremost principle of web accessibility. Visually impaired users using screen readers will use these textual descriptions to get better understanding of images present in digital contents. In this paper, we propose Chart-Text a novel fully automated system that creates textual description of chart images. Given a PNG image of a chart, our Chart-Text system creates a complete textual description of it. First, the system classifies the type of chart and then it detects and classifies the labels and texts in the charts. Finally, it uses specific image processing algorithms to extract relevant information from the chart images. Our proposed system achieves an accuracy of 99.72% in classifying the charts and an accuracy of 78.9% in extracting the data and creating the corresponding textual description.

**Keywords:** Computer vision, Chart data extraction, Information retrieval, Object detection

**1 Introduction**

Images are an excellent medium for data representation and interpretation. For the images present in digital contents to be useful to people with visual challenges, it is mandatory for the content creators to provide a corresponding textual description. This process of providing textual description for an image is termed as Alternative Text (Alt-Text). These Alt-Texts are present as part of E-books and other similar digital contents that are read aloud to users with visual challenges using screen readers. This process of creating textual description for images is a highly time consuming manual process. Moreover, for the same image the textual description varies from person to person. Thus, standardizing these Alt-Texts poses a challenge. Hence, having a system that can create textual description of the images will boost the credibility and the productivity of the content creators. It also enhances the accessibility of the digital content to people with visual challenges. To address this, we developed Chart-Text, a system that can create textual description given a chart image.

Our proposed Chart-Text system creates complete textual description of 5 different type of chart images namely: pie charts, horizontal bar charts, vertical bar charts, stacked horizontal bar charts and stacked vertical bar charts. Our system achieves an accuracy of 99.72% in classifying the charts and an accuracy of 78.9% in extracting the data and creating the corresponding Alt-Text. The paper is organized as follows: First, we discuss a few related works in section 2. Then, we describe the dataset that we used to create the system in section 3. In section 4, we provide the proposed methodology for extracting information from chart images. Then, we provide the results along with sample input and outputs.

## 2 Related works

We built our system by leveraging past research works in areas such as image classification, object detection, localization and optical character recognition. A few different systems have been built using the above approaches. Most of these systems are semi-automated except for [1].

ReVision [2] is one of the initial systems developed for automatically redesigning charts to improve graphic depictions to help users in understanding the data better. It first classifies the type of chart and then uses traditional image processing algorithms to extract the marks and the data from the charts. Then, ReVision applies heuristics to map the marks and the related data extracted from the graphs. Finally, the system combines the existing information extracted from the charts and comes up with an easily decipherable visualization.

ChartSense [3] is a system similar to ReVision but has a better performance in classifying the chart type. This increase in performance is attributed to the deep learning based image classifier as compared to the traditional image processing based classifier in ReVision. ChartSense overall is a semi-automatic system that uses input from users to further improve detection accuracy. ChartSense applies an interactive data extraction algorithm most appropriate for the identified chart type. In addition, ChartSense provides a set of simple interactions to fine-tune the result, enabling more accurate data extraction.

Financial analysis includes many visualizations and scatter plots are widely used. To further facilitate data inspections, analysis and financial modeling, it is necessary to extract information from the scatter plots. Scatteract [1] is a fully automated solution developed to achieve the purpose of performing scatter plot data extraction. It combines deep learning and image-processing algorithms to identify the scatterplot components and OCR to extract the text information from the scatterplots.

## 3 Dataset

Chart-Text uses both image classification and object detection models, which require a large number of annotated images to train. To the best of our knowledge, there is no publicly available dataset that satisfies our needs. We chose to follow the method given in [1] and generated images using Matplotlib [4]. We generated 5000 training and 1000 testing images for each of pie charts, horizontal bar charts, vertical bar charts, stacked horizontal and stacked vertical bar charts. We used the same set of training images for the classifier and the object detector. Thus, we have five classes for classification: horizontal bar charts, vertical bar charts, stacked horizontal bar charts and stacked vertical bar charts and seven classes for object detection as given in Table 1. We randomized the plot aesthetics as given below. This is adopted from [1].

Plot aesthetics:
- Plot style (default styling values e.g. "classic", "seaborn", "ggplot", etc.)
- Size and resolution of the plot
- Number of style variations on the points
- Point styles (markers, sizes, colors)
- Padding between axis and tick values, axis and axis-labels, and plot area and plot title
- Axis and tick's locations
- Tick sizes if using non-default ticks
- Rotation angle of the tick values, if any
- Presence and style of minor ticks
- Presence and style of grid lines
- Font and size of the tick values
- Fonts, sizes and contents of the axis-labels and plot title
- Colors of the plot area and background area

**Table 1**. Labels present in the procedurally generated dataset for text position detection.

| Class | Presence in charts |
|---|---|
| Chart title | All charts |
| X value | Except pie charts |
| Y value | Except pie charts |
| X label | Except pie charts |
| Y label | Except pie charts |
| Label | Only pie charts |
| Legend | Randomly present in pie charts and definitely present in stacked horizontal and stacked vertical bar charts |

**4 Chart-Text system:**

Chart-Text system comprises a deep learning based classifier and a deep learning based object detector. The classifier identifies the type of the input chart and the object detector detects the various text regions in the charts. Optical character recognition is applied to extract the texts from the bounding boxes identified by the object detector. Further, the system extracts the relevant data from the charts using chart specific image processing algorithms. Finally, all extracted information is sent as input to predetermined text templates (one template per type of chart) to generate the final Alt-Text. We address each of the above steps in detail in the following sections.

**4.1 Charts classification:**

Recently, Deep Neural Networks (DNN), particularly Convolutional Neural Networks (CNNs) have excelled in recognizing objects in the images [5]. CNNs are a type of neural network architecture specially made to deal with image data. CNNs typically consist of Convolutional Layer, Max Pooling layer and Fully Connected layers. Convolution layer tangles nearby pixels to abstract their meaning. Pooling layer extracts representative specimens from the result of the convolution layer to reduce computational time. The fully connected layer does conventional neural network learning. Of different CNN architectures available publicly, we choose MobileNet [6] as our CNN architecture. They achieve a similar accuracy to VGG-16 [7] using far fewer parameters on ImageNet [8]. MobileNet achieves 70.6% accuracy using only 4.6 million parameters compared to VGG-16's 71.5% using 138 million parameters.

**4.1.1 Training the classifier:**

We used Keras [9] with Tensorflow [10] backend to train the classification network on our data. We resized the images to a fixed size of 224 x 224 x 3 and normalized them before feeding into the network. We used transfer learning [11] by initializing the network with ImageNet weights. We used SGD as our optimizer with Nesterov momentum and trained it for 50 epochs. The learning rate (LR) schedule we used is as follows: we start with a LR of 0.0001 and reduce it by a factor of 30% after every 20 epochs. We used a mini-batch size of 64 and the network was trained on Nvidia Tesla M60 (2 GPUs with 8GB memory each) for approximately 3.5 hours. The performance of the classifier is discussed in the results section.

**4.2 Text position detection**

Detecting objects in an image is a common problem in computer vision. The task involves putting bounding boxes around objects of interest in the image. In Chart-Text we use object detection to localize and classify chart title, x-label, y-label, x-values, y-values, labels and legends as described in the dataset section. There are various object detection systems but, all state of the art object detection systems uses deep learning. We use Faster R-CNN [12] as our object detection model because, as [13] argues, this achieves the best accuracy for small object detection when compared to YOLO[14], SSD[15] and R-FCN[16] with the trade-off being the inference speed.

### 4.2.1 Training the object detector

We used Tensorflow's Object detection API to train the Faster R-CNN object detection model. We used Inception-V2[17] as our fixed feature extractor. We resized all the incoming images while preserving their aspect ratio in such a way that their shorter side is at least 500 pixels. We trained the network for 50,000 iterations with a mini-batch size of 1. We used the following anchor box sizes: [0.25, 0.5, 0.75, 1.0, 1.5] (w.r.t to a base size of 256 pixels) and the following aspect ratios: [0.5, 1.0, 2.0]. The total training time was about 4 hours on an Nvidia Tesla M60. After training, given a chart image, the model outputs 300 bounding boxes with confidence scores and class names. We only keep the detections with a confidence score greater than or equal to 0.5. We chose this number based on validation experiments.

### 4.3 Extracting text from charts

To extract text information from the charts we apply Optical character recognition (OCR) to the bounding boxes output from our object detection model. We use Google's open source Tesseract [18] as our OCR engine. Out of the box, it comes pre-trained to recognize several fonts and styles. It is possible to retrain it on custom fonts but we use it as it is. As pointed out in [1], the OCR accuracy is higher for horizontally aligned text. We use the following algorithm to align the text image horizontally:

1. We convert the image to grey scale and threshold it to obtain a binary image.
2. We fit a minimum bounding rectangle to the binary image and measure the angle made by the rectangle w.r.t. the horizontal.
3. We rotate the image using the calculated angle in the opposite direction to align the text image with the horizontal.

We also found that rescaling the aligned image so that its height is 130 pixels significantly improves OCR accuracy. The above procedure is very similar to the one given in [1]. Figure 1 shows the results of the alignment

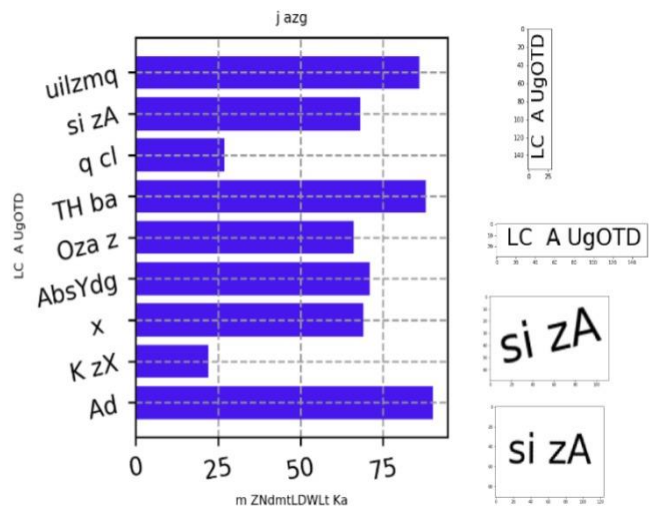

**Figure 1**: example of horizontally aligning text images to facilitate OCR using Tesseract OCR engine.

### 4.4 Extracting data from charts

In this section, we present the chart specific image processing algorithms that we use to extract relevant data marks from the chart images. As the data extraction process is unique for every type of chart, we first provide the underlying assumptions and then describe the specific image processing algorithm.

### 4.4.1 Horizontal and vertical bar charts

We assumed that horizontal and vertical bar charts do not contain 3D effects. The algorithm

for horizontal and vertical bar extraction is as follows:
1. We convert the image to grayscale and then use a median filter with a kernel size of 7 to remove noises.
2. We threshold the image and perform area opening morphological operation to remove any remaining small objects.
3. We label the regions, perform connected components analysis and fit a minimum bounding rectangle to each of the labelled region to identify and extract different bars.
4. We use OpenCV [19] and DIPlib [20] to perform the above-mentioned operations.

Here, we do not consider the chart orientation (horizontal or vertical) as this is taken care of by the classifier, and this method works irrespective of the chart orientation.

### 4.4.2 Stacked horizontal and vertical bar charts

Similar to horizontal and vertical bar charts, we assume that the charts do not have 3D effects. Additionally, we also assume that there are no texts inside the stacked bars and each stack has a distinct color. We follow the same procedure as horizontal and vertical bar charts to extract different bars. In addition, after extracting the individual bars we use a local search algorithm to extract the stacks. We confine our search space to the bars extracted from the connected component analysis. The local search algorithm is as follows:
1. We find the midpoint of the two edges of the rectangle along its length. For horizontal bar charts lengthwise is the x-axis and for vertical bar charts lengthwise is the y-axis.
2. With these two points as start and end points, we move along the line made by them and detect sudden changes in pixel values between two adjacent pixel points. We take the center of those pixel points as the border or the edge point separating two adjacent stacks.
3. The process continues until we find all the stacks inside all the bars.

For all the 4 types of bar charts we calculate a ratio called as *chart to pixel ratio*. We calculate this ratio by extracting the numbers and position of y values (horizontal charts) or x values (vertical charts). We define the *chart to pixel ratio* as the ratio of two consecutive x or y values extracted using OCR and the corresponding distance between the centroids of their bounding boxes. We use this ratio to scale the bars to the chart coordinate system from the pixel coordinate system.

### 4.4.3 Pie chart
We assume pie charts do not have 3D effects, each wedge has a distinct colour, and the corresponding labels are present adjacent to the wedge. We adopted a similar approach to the one given in [3]. The approach is explained in detail below:
1. We use Canny edge detection to create an edged image of the pie chart.
2. We fit a robust circle to the pie using RANSAC. We use RANSAC because it is more efficient in ignoring outliers while fitting a circle.
3. We sample 1000 pixels uniformly along the fitted circle whose radius is 0.2*R. Where, R is the radius of the fitted circle.
4. We traverse the sampled points sequentially and detect sudden change in pixel intensity between two adjacent pixel points. We take the center of those two pixel points as the border point between wedges.
5. After detecting the border points, we measure the clockwise angle between them with respect to the center of the fitted circle to determine the angle of each wedge.

## 5 Results

### 5.1 Classifier and Detector results

Our Chart-Text system achieved a classification accuracy of 99.72% on the procedurally generated test set. The normalized confusion matrix for classification is given in Table 2. It is evident from the confusion matrix that the classifier misclassified three stacked vertical bar chart images as vertical bar chart and four horizontal bar chart images as stacked horizontal bar charts and six stacked horizontal bar chart images as horizontal bar chart.

Understanding where the CNN is looking in the image to classify is very important to understand the generalizing capability of the network. Grad-Cam [21] visualizations give insights on which part of the image maximally activates the neurons to predict that particular class. We generated Grad-Cam images to verify whether the network attends to correct part of the image when it predicts a particular class. Figure 2 shows a few samples of Grad Cam results. These images show that the network attends to the correct part of the image while classifying.

Similarly, we evaluate the performance of the text position detection (object detection) model on the procedurally generated test set. Generally, the performance of an object detector is measured by the mean-average-precision(mAP) metric. The mAP for each class at an Intersection over Union (IOU) of 50% is given in Table 3

**Table 2.** Normalized Confusion Matrix

| Predicted Label<br>True Label | Horizontal bar chart | Pie chart | Stacked horizontal bar chart | Stacked vertical bar chart | Vertical bar chart |
|---|---|---|---|---|---|
| Horizontal bar chart | 0.996 | 0.0 | 0.004 | 0.0 | 0.0 |
| Pie chart | 0.0 | 1.0 | 0.0 | 0.0 | 0.0 |
| Stacked horizontal bar chart | 0.006 | 0.0 | 0.994 | 0.0 | 0.0 |
| Stacked vertical bar chart | 0.001 | 0.0 | 0.0 | 0.996 | 0.003 |
| Vertical bar chart | 0.0 | 0.0 | 0.0 | 0.0 | 1.0 |

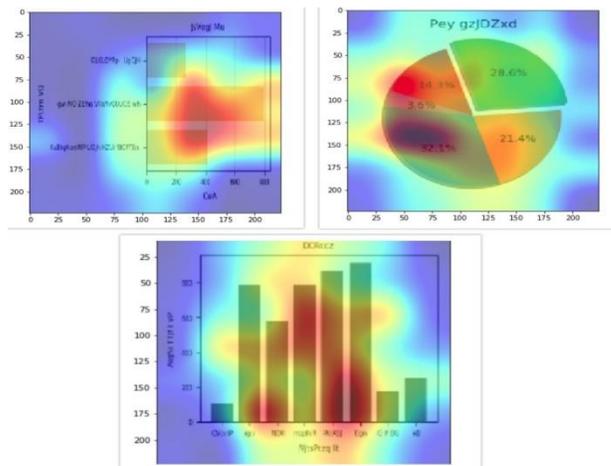

**Figure 2**: Grad-Cam images (clockwise) first image, depicts the pixel regions which maximally activate the class 'horizontal bar chart'. Second image, depicts the pixel regions which maximally activate the class 'pie chart'. Third image depicts the pixel regions which maximally activate the class 'vertical bar chart'.

Table 3: Mean Average Precision metric for the Object detection model

| Class | mAP at 0.5IOU |
|---|---|
| Chart title | 0.9796 |
| Pie Label | 0.9304 |
| Legend | 0.8857 |
| X label | 0.9486 |
| X value | 0.9365 |
| Y label | 0.9478 |
| Y value | 0.9313 |

## 5.2 Alt-Text accuracy

The performance of Alt-Text generation is dependent on accuracy of the extracted text from the charts and the precision of the calculated values from the chart. To evaluate the performance of the complete system in generating the Alt-Text on the procedurally generated plots we define two evaluation metrics.

**Text accuracy**: We define that the extracted text is true positive if the Levenshtein distance ratio between the extracted text and the ground truth text is greater than 80%. We use 80% instead of a 100% match because, the texts are randomly generated and in some cases have special characters in them. Mathematically:

$$\frac{Dist(P,T)}{Alignment\ distance} \geq 0.8 \quad (1)$$

Where, *Dist* is Levenshtein distance, P is the predicted string and T is ground truth string.

**Value accuracy**: we define that the calculated value is true positive if and only if the difference between the calculated value and ground truth-value is within 2%. Mathematically:

$$\frac{|Calculated\ value - True\ value|}{True\ value} \leq 0.2 \quad (2)$$

For each image belonging to a particular class, we calculate the mean text and value accuracies and average them across all the images of that particular class and report this average text and value accuracies for each of the five classes in Table 4.

Table 4. Average text and value accuracies of all the 5 types of charts

| Class | Average text accuracy (%) | Average value accuracy (%) |
|---|---|---|
| Pie chart | 85 | 80.7 |
| Horizontal bar chart | 82.3 | 76.4 |
| Vertical bar chart | 83.5 | 76.5 |
| Stacked horizontal bar chart | 80.47 | 71.64 |
| Stacked vertical bar chart | 81.2 | 70.12 |

## 6 Conclusion

In this paper, we introduced Chart-Text, a fully automated system that completely describes horizontal bar charts, vertical bar charts, stacked horizontal bar charts, stacked vertical bar charts and pie charts. We developed an original approach to extract data from stacked bar charts. One more important contribution in Chart-Text is our deep learning based object detection model that is used to detect and classify chart title, pie labels, x values, y values, x labels, and y labels. Our method is generic and can be extended it to other chart types such as line charts, scatter plots, area charts, histograms etc.

Even though our system performs considerably well on the procedurally generated dataset, the performance on data from web and other sources plummets. This is primarily because most charts from web and other sources are graphically enhanced and typesetted for better visual appearance. Further investigations on improving Chart-Text's generalization capability is needed. The ultimate goal of Chart-Text is to act as an end-to-end fully automatic system that extracts information from charts to create complete textual description of it so that the chart images become more accessible to people with vision impairments.

**7 Examples**

Here, we provide a few Alt-Texts generated by our Chart-Text system. These example input images are generated using Matplotlib.

**Stacked Vertical Bar Chart**

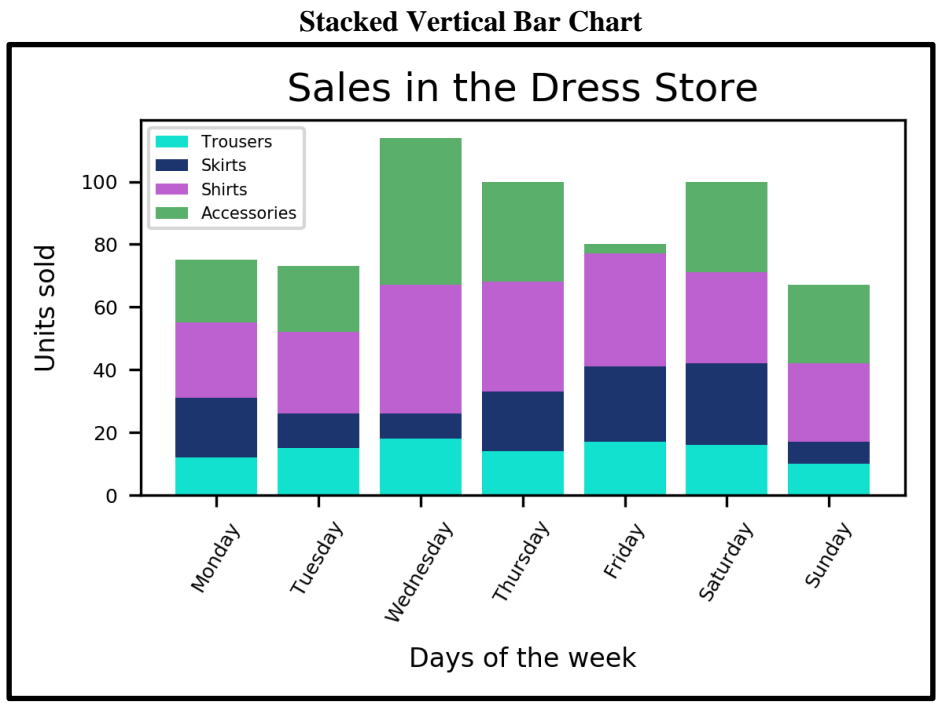

**Alt-Text:** The stacked vertical bar graph depicts 'Sales in the Dress Store'. The graph is plot between 'Units sold' y axis over 'Days of the week' x axis for '(Shirts, Skirts, Trousers, Accessories)'. The Units sold for the corresponding Days of the week are (Monday Accessories = 20, Shirts = 24, Skirts = 19, Trousers = 13), (Thursday Accessories = 31, Shirts = 35, Skirts = 19, Trousers = 15), (Saturday Accessories = 28, Shirts = 29, Skirts = 26, Trousers = 17), (Friday Accessories = 3, Shirts = 36, Skirts = 24, Trousers = 18), (Tuesday Accessories = 21, Shirts = 26, Skirts = 11, Trousers = 16), (Wednesday Accessories = 46, Shirts = 41, Skirts = 8, Trousers = 19), (Sunday Accessories = 25, Shirts = 25, Skirts = 7, Trousers = 11). (All Values estimated)

**Stacked Horizontal Bar Chart**

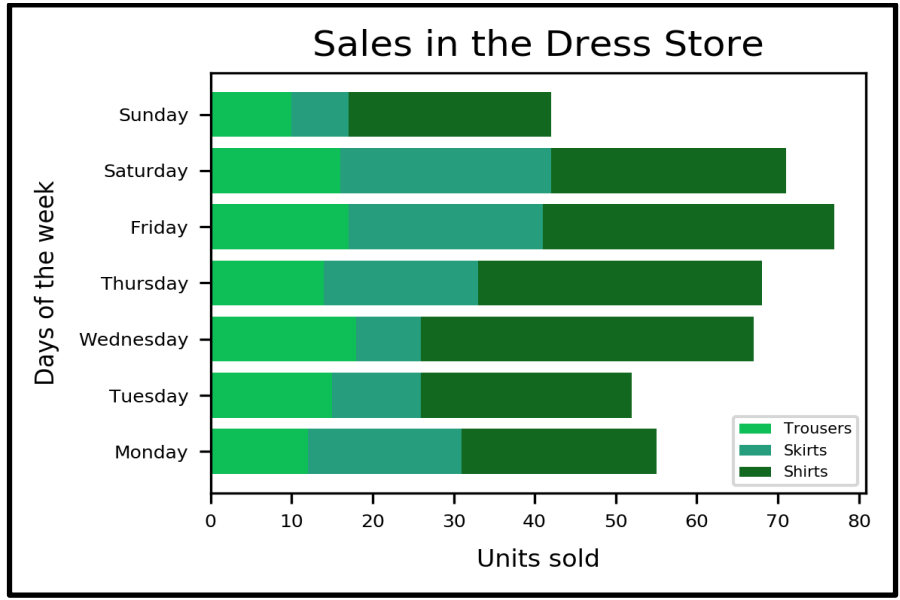

**Alt-Text:** The stacked horizontal bar graph depicts 'Sales in the Dress Store'. The graph is plot between 'Units sold' x axis over 'Days of the week' y axis for '(Shirts, Trousers, Skirts)'. The Units sold for the corresponding Days of the week are (Monday Trousers = 11, Skirts = 17, Shirts = 21), (Thursday Trousers = 12, Skirts = 17, Shirts = 31), (Saturday Trousers = 14, Skirts = 23, Shirts = 25), (Friday Trousers = 15, Skirts = 21, Shirts = 32), (Tuesday Trousers = 13, Skirts = 10, Shirts = 23), (Wednesday Trousers = 16, Skirts = 7, Shirts = 36), (Sunday Trousers = 9, Skirts = 6, Shirts = 22). (All Values estimated)

**Vertical Bar Chart**

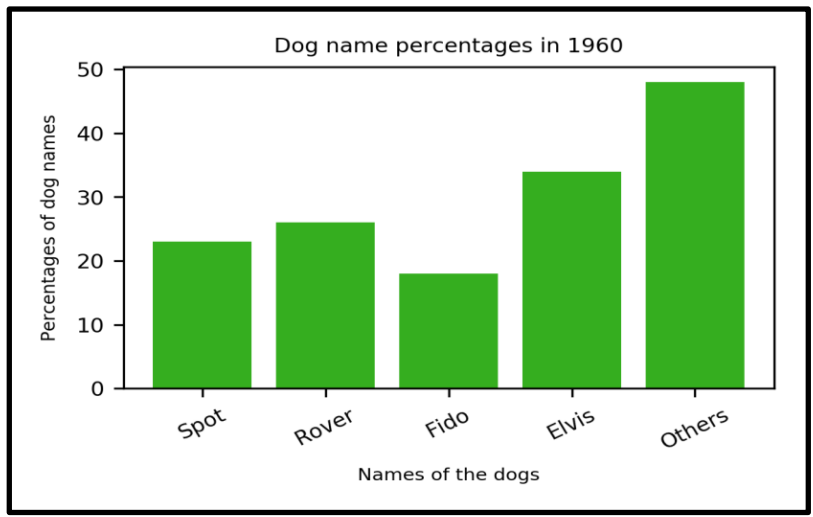

**Alt-Text:** The vertical bar graph depicts 'Dog name percentages in 1960'. The graph is plot between 'Percentages of dog names' y axis over 'Names of the dogs' x axis. The Percentages of dog names for the corresponding Names of the dogs are Others = 48, Elvis = 34, Rover = 26, Spot = 23, Fido = 18. (All Values estimated)

**Horizontal Bar Chart**

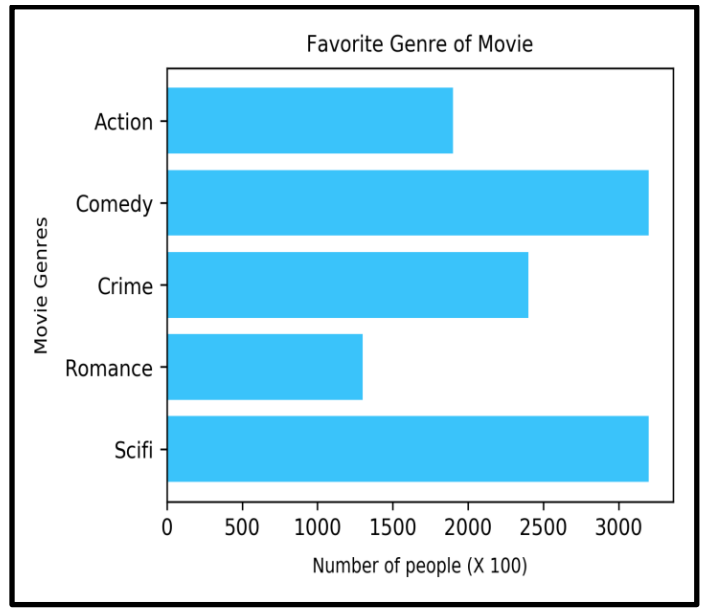

**Alt-Text:** The horizontal bar graph depicts 'Favorite Genre of Movie'. The graph is plot between 'Number of people (X 100)' x-axis over 'Movie Genres' y-axis. The Number of people (X 100) for the corresponding Movie Genres are Action = 1896, Comedy = 3187, Crime = 2393, Romance = 1300, Scifi = 3187. (All Values estimated)

**Pie Chart**

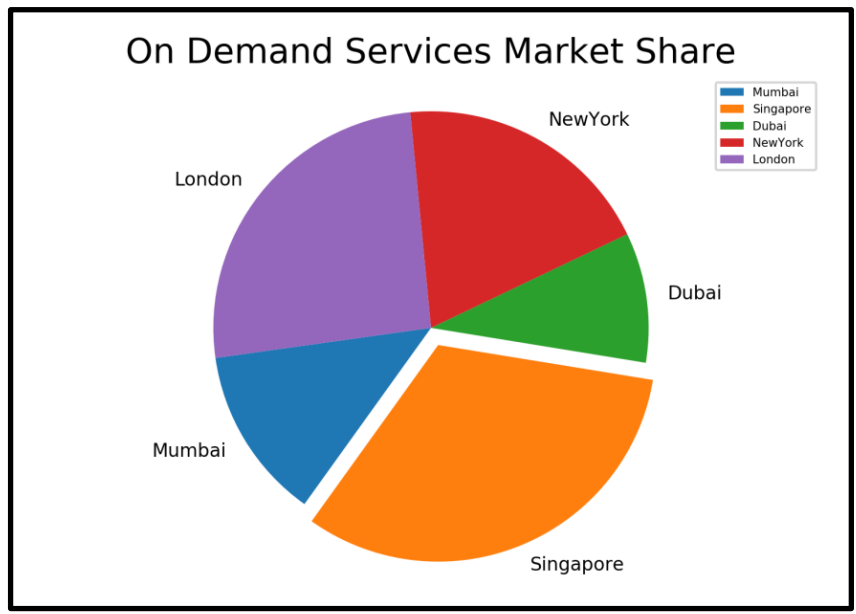

**Alt-Text:** The pie chart depicts 'On Demand Services Market Share'.It has 5 parts (Dubal, New York, London, Mumbai, Singapore) to it and percentage for each are Dubal: 11%, New York: 18%, London: 27%, Mumbai: 13%, Singapore: 32%. (All Values estimated)

**Stacked Vertical Bar Chart**

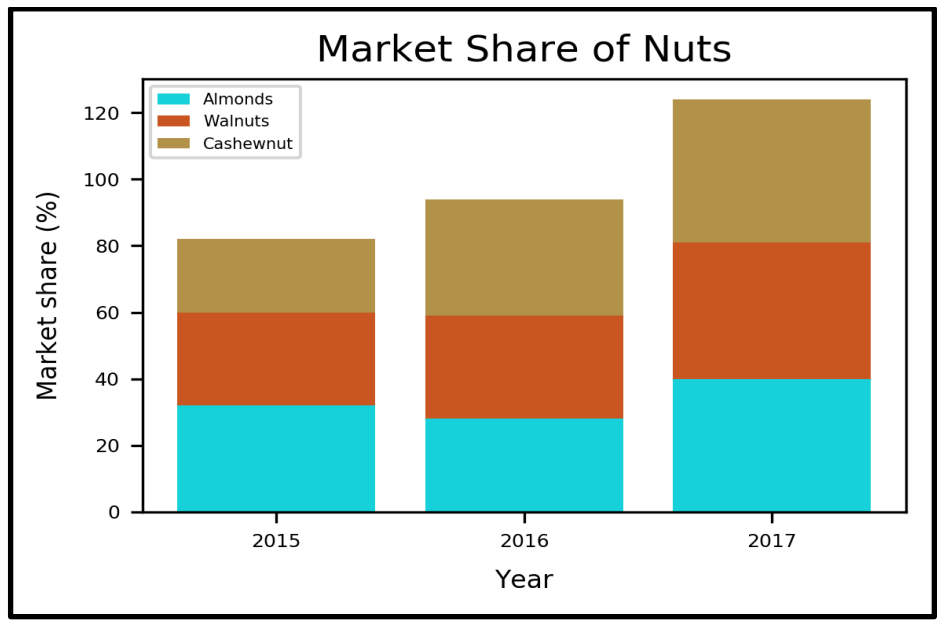

**Alt-Text:** The stacked vertical bar graph depicts 'Market Share of Nuts'. The graph is plot between 'Market share (%)' y axis over 'Year' x axis for '(Cashewnut, Walnuts, Almonds)'. The Market share (%) for the corresponding Year are (2015 Cashewnut = 22, Walnuts = 28, Almonds = 33), (2017 Cashewnut = 43, Walnuts = 41, Almonds = 41), (2016 Cashewnut = 65, Walnuts = 29). (All Values estimated)